\documentclass[10pt,twocolumn,letterpaper]{article}

\usepackage{cvpr}
\usepackage{times}
\usepackage{epsfig}
\usepackage{graphicx}
\usepackage{amsmath}
\usepackage{amssymb}
 \makeatletter
 \DeclareRobustCommand\onedot{\futurelet\@let@token\@onedot}
 \def\@onedot{\ifx\@let@token.\else.\null\fi\xspace}

 \makeatother

\usepackage{subcaption}
\usepackage{soul}
\newcommand{\mysubsubsection}[1]{\textbf{#1.}}

\usepackage[breaklinks=true,bookmarks=false]{hyperref}
\cvprfinalcopy 

\def\cvprPaperID{788} 

\ifcvprfinal\fi
\begin{document}

\title{End-to-end Learning of Driving Models from Large-scale Video Datasets}

\author{Huazhe Xu$^{*}$ \quad Yang Gao$^{*}$ \quad Fisher Yu \quad Trevor Darrell\\
University of California, Berkeley}

\maketitle
\thispagestyle{empty}

\begin{abstract}

Robust perception-action models should be learned from training data with diverse visual appearances and realistic behaviors, yet current approaches to deep visuomotor policy learning have been generally limited to \textit{in-situ} models learned from a single vehicle or simulation environment. We advocate learning a generic vehicle motion model from large scale crowd-sourced video data, and develop an end-to-end trainable architecture for learning to predict a distribution over future vehicle egomotion from instantaneous monocular camera observations and previous vehicle state. Our model incorporates a novel FCN-LSTM architecture, which can be learned from large-scale crowd-sourced vehicle action data, and leverages available scene segmentation side tasks to improve performance under a privileged learning paradigm. We provide a novel large-scale dataset of crowd-sourced driving behavior suitable for training our model, and report results predicting the driver action on held out sequences 
across diverse conditions.  
\stepcounter{footnote}
\footnotetext[1]{* indicates equal contribution}

\end{abstract}
 
\section{Introduction}

Learning perception-based policies to support complex autonomous behaviors, including driving, is an ongoing challenge for computer vision and machine learning.  While recent advances that use rule-based methods have achieved some success, we believe that learning-based approaches will be ultimately needed to handle complex or rare scenarios, and scenarios that involve multi-agent interplay with other human agents.

The recent success of deep learning methods for visual perception tasks has increased interest in their efficacy for learning action policies.  Recent demonstration systems \cite{bojarski2016end,chen2015deepdriving,lecun2005off} have shown that simple tasks, such as a vehicle lane-following policy or obstacle avoidance, can be solved by a neural net. This echoes the seminal work by Dean Pomerleau with the CMU NavLab, whose ALVINN network was among the earliest successful neural network models \cite{Pomerleau__1989_2055}.

These prior efforts generally formulate the problem as learning a mapping from pixels to actuation.  This end-to-end optimization is appealing as it directly mimics the demonstrated performance, but is limiting in that it can only be performed on data collected with the specifically calibrated actuation setup, or in corresponding simulations (e.g., as was done in \cite{Pomerleau__1989_2055}, and more recently in~\cite{Tzeng16WAFR, rusu2016sim, Daftry2016ISER}).  The success of supervised robot learning-based methods is governed by the availability of training data, and typical publicly available datasets only contain on the order of dozens to hundreds of hours of collected experience. 

\begin{figure}[t]
\begin{center}
   \includegraphics[width=\linewidth]{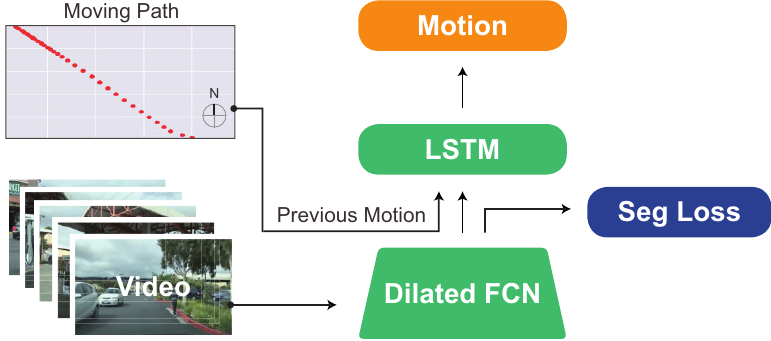} 
\end{center}
   \caption{Autonomous driving is formulated as a future egomotion prediction problem. Given a large-scale driving video dataset, an end-to-end FCN-LSTM network is trained to predict multi-modal discrete and continuous driving behaviors. Using semantic segmentation as a side task further improves the model.}
\label{fig:concept}
\end{figure}

We explore an alternative paradigm, which follows the successful practice in most computer vision settings, of exploiting large scale online and/or crowdsourced datasets.  We advocate learning a driving model or policy from large scale uncalibrated sources, and specifically optimize models based on crowdsourced dashcam video sources. We release with our paper a curated dataset from which suitable models or policies can be learned.

To learn a model from this data, we propose a novel deep learning architecture for learning-to-drive from uncalibrated large-scale video data.  We formulate the problem as learning a generic driving model/policy; our learned model is generic in that it learns a predictive future motion path given the present agent state.  Presently we learn our model
from a corpus of demonstrated behavior and evaluate on held out data from the same corpus.
Our driving model is akin to a language model, which scores the likelihood of character or word sequences given certain corpora; our model similarly is trained and evaluated in terms of its ability to score as highly likely the observed behavior of the held out driving sequence.  It is also a policy in that it defines a probability distribution over actions conditioned on a state, with the limitation that the policy is never actually executed in the real world or simulation.

Our paper offers four novel contributions. First, we introduce a generic motion approach to learning a deep visuomotor action policy where actuator independent motion plans are learned based on current visual observations and previous vehicle state.  
Second, we develop a novel FCN-LSTM which can learn jointly from demonstration loss and segmentation loss, and can output multimodal predictions. Third, we curate and make publicly available a large-scale dataset to learn a generic motion model from vehicles with heterogeneous actuators. Finally, we report experimental results confirming that ``privileged" training with side task (semantic segmentation) loss learns egomotion prediction tasks faster than from motion prediction task loss alone\footnote{The codebase and dataset can be found at \url{https://github.com/gy20073/BDD_Driving_Model/}}.

We evaluate our model and compare to various baselines in terms of the ability of the model to predict held-out video examples; our task can be thought of that of predicting future egomotion given present observation and previous agent state history.

While future work includes extending our model to drive a real car, and addressing issues therein involving policy coverage across undemonstrated regions of the policy space (c.f. \cite{ross2011reduction}), we nonetheless believe that effective driving models learned from large scale datasets using the class of methods we propose will be a key element in learning a robust policy for a future driving agent.

\section{Related Work}

ALVINN \cite{Pomerleau__1989_2055} was among the very first attempts to use a neural network for autonomous vehicle navigation. 
The approach was simple, comprised of a shallow network that predicted actions from pixel inputs applied to simple driving scenarios with few obstacles; nevertheless, its success suggested the potential of neural networks for autonomous navigation. 

Recently, NVIDIA demonstrated a similar idea that benefited from the power of modern convolution networks to extract features from the driving frames \cite{bojarski2016end}.
This framework was successful in relatively simple real-world scenarios, such as highway lane-following and driving in flat, obstacle-free courses.

Instead of directly learning to map from pixels to actuation, \cite{chen2015deepdriving} proposed mapping pixels to pre-defined affordance measures, such as the distance to surrounding cars.
This approach provides human-interpretable intermediate outputs, but a complete set of such measures may be intractable to define in complex, real-world scenarios.
Moreover, the learned affordances need to be manually associated with car actions, which is expensive, as was the case with older rule-based systems.
Concurrent approaches in industry have used neural network predictions from tasks such as object detection and lane segmentation as inputs to a rule-based control system \cite{huval2015empirical}.

Another line of work has treated autonomous navigation as a visual prediction task in which future video frames are predicted on the basis of previous frames.
\cite{santana2016learning} propose to learn a driving simulator with an approach that combines a Variational Auto-encoder (VAE) \cite{kingma2014auto} and a Generative Adversarial Network (GAN) \cite{goodfellow2014generative}.
This method is a special case of the more general task of video prediction; there are examples of video prediction models being applied to driving scenarios \cite{de2016dynamic, lotter2016deep}. However, in many scenarios, video prediction is ill-constrained as preceding actions are not given as input the model. \cite{oh2015action, finn2016unsupervised} address this by conditioning the prediction on the model's previous actions.
In our work, we incorporate information about previous actions in the form of an accumulated hidden state.

Our model also includes a side- or privileged-information learning aspect. This occurs when a learning algorithm has additional knowledge at training time; i.e., additional labels or meta-data. This extra information helps training of a better model than possible using only the view available at test time. A theoretical framework for learning under privileged information (LUPI) was explored in~\cite{vapnik2009new}; a max-margin framework for learning with side-information in the form of bounding boxes, image tags, and attributes was examined in \cite{sharmanska2013learning} within the DPM framework. Recently \cite{Hoffman_CVPR2016} exploited deep learning with side tasks when mapping from depth to intensity data. Below we exploit a privileged/side-training paradigm for learning to drive, using semantic segmentation side labels.

Recent advances in recurrent neural network modeling for sequential image data are also related to our work. The Long-term Recurrent Convolutional Network (LRCN) \cite{donahue2015long} model investigates the use of deep visual features for sequence modeling tasks by applying a long short-term memory (LSTM) recurrent neural network to the output of a convolutional neural network. We take this approach, but use the novel combination of a fully-convolutional network (FCN) \cite{long2015fully} and an LSTM. A different approach is taken by \cite{xingjian2015convolutional}, as they introduce a convolutional long short-term memory (LSTM) network that directly incorporates convolution operations into the cell updates.

\section{Deep Generic Driving Networks}
We first describe our overall approach for learning a generic driving model from large-scale driving behavior datasets, and then propose a specific novel architecture for learning a deep driving network.

\subsection{Generic Driving Models}

We propose to learn a generic approach to learning a driving policy from demonstrated behaviors, and formulate the problem as predicting future feasible actions.  Our  driving model is defined as the admissibility of which next motion  is  plausible given the current observed world configuration. Note that the world configuration incorporates previous observation and vehicle state. 
Formally, a driving model $F$ is a function defined as:
\begin{equation}
F(s, a): S\times A \rightarrow \mathbb{R}
\end{equation}
where $s$ denotes states, $a$ represents a potential motion action and $F(s, a)$ measures the feasibility score of operating motion action $a$ under the state $s$. 

Our approach is {\it generic} in that it predicts egomotion, rather than actuation of a specific vehicle.\footnote{Future work will comprise how to take such a prediction and cause the desired motion to occur on a specific actuation platform. The latter problem has been long studied in the robotics and control literature and both conventional and deep-learning based solutions are feasible (as is their combination). }
Our generic models take as input raw pixels and current and prior vehicle state signals, and predict the likelihood of future motion.  This can be defined over a range of action or motion granularity, and we consider both discrete and continuous settings in this paper.\footnote{We leave the most general setting, of predicting directly arbitrary 6DOF motion, also to future work.}  
For example, the motion action set $A$ could be a set of coarse actions: 
\begin{equation}
A = \{\textrm{straight, stop, left-turn, right-turn}\}
\label{act:discrete}
\end{equation}
One can also define finer actions based on the car egomotion heading in the future. In that case, the possible motion action set is:
\begin{equation}
A = \{ \vec{v} | \vec{v} \in \mathbb{R}^2\}
\label{act:continuous}
\end{equation}
where, $\vec{v}$ denotes the future egomotion on the ground plane. 

We refer to $F(s,a)$ as a driving model inspired by its similarity to the classical N-gram language model in Natural Language Processing. Both of them take in the sequence of prior events, such as what the driver has seen in the driving model, or the previously observed tokens in the language model, and predict plausible future events, such as the viable physical actions or the coherent words. Our driving model can equivalently be thought of as a policy from a robotics perspective, but we presently only train and test our model from fixed existing datasets, as explained below, and consequently we feel the language model analogy is the more suitable one.

\begin{figure}
\begin{center}
\includegraphics[width=\linewidth]{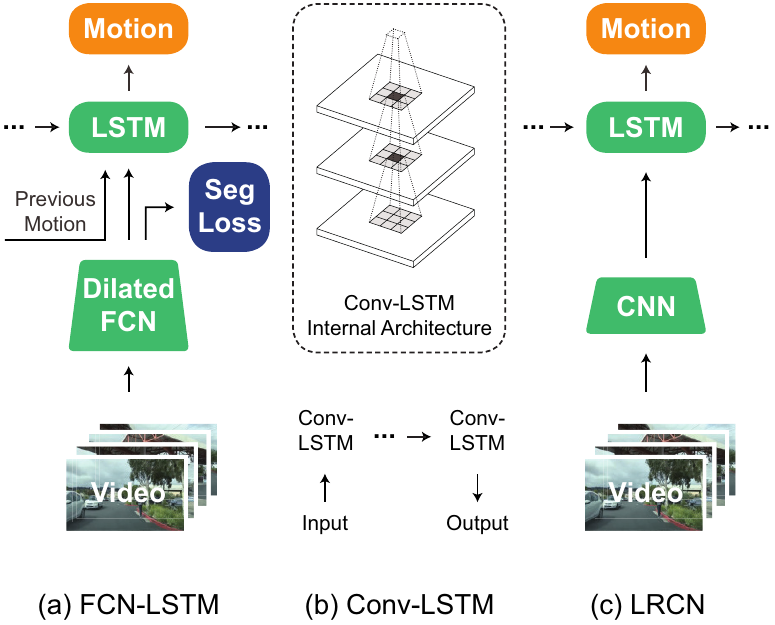}
\end{center}
\caption{Comparison among novel architectures that can fuse time-series information with visual inputs.}
\label{fig:arch}
\end{figure}

\subsection{FCN-LSTM Architecture}
\label{sec:arch}

Our goal is to predict the distribution over feasible future actions, conditioned on the past and current states, including visual cues and egomotions. To accomplish our goal, an image encoder is necessary to learn the relevant visual representation in each input frame, together with a temporal network to take advantage of the motion history  information.  We propose a novel architecture for time-series prediction which fuses an LSTM temporal encoder with a fully convolutional visual encoder.  Our model is able to jointly train motion prediction and pixel-level supervised tasks. We can use semantic segmentation as a side task following ``previleged'' information learning paradigm. This leads to better performance in our experiments. Figure \ref{fig:arch} compares our architecture (FCN-LSTM) with two related architectures\cite{donahue2015long, xingjian2015convolutional}.

\subsubsection{Visual Encoder}
Given a video frame input, a visual encoder can encode the visual information in a discriminative manner while maintaining the relevant spatial information. In our architecture, a dilated fully convolutional neural network \cite{yu2015multi, donahue2015long} is used to extract the visual representations. We take the ImageNet \cite{russakovsky2015imagenet} pre-trained AlexNet \cite{krizhevsky2012imagenet} model, remove POOL2 and POOL5 layers and use dilated convolutions for conv3 through fc7. To get a more discriminative encoder, we finetune it jointly with the temporal network described below. The dilated FCN representation has the advantage that it enables the network to be jointly trained with a side task in an end-to-end manner. This approach is advantageous when the training data is scarce.

\subsubsection{Temporal Fusion}
We optionally concatenate the past ground truth sensor information, such as speed and angular velocity, with the extracted visual representation. With the visual and sensor states at each time step, we use an LSTM to fuse all past and current states into a single state, corresponding to the state $s$ in our driving model $F(s, a)$. This state is complete, in the sense that it contains all historical information about all sensors. We could predict the physical viability from the state $s$ using a fully connected layer. 

We also investigate below another temporal fusion approach, temporal convolution, instead of LSTM to fuse the temporal information. A temporal convolution layer takes in multiple visual representations and convolves on the time dimension with an $n \times 1$ kernel where n is the number of input representations. 

\subsection{Driving Perplexity }
Our goal is to learn a future motion action feasibility distribution, also known as the driving model. However, in past work \cite{Pomerleau__1989_2055,chen2015deepdriving,bojarski2016end}, there are few explicit quantitative evaluation metrics. In this section, we define an evaluation metrics suitable for large-scale uncalibrated training, based on sequence perplexity.

Inspired by language modeling metrics, we propose to use perplexity as evaluation metric to drive training. For example, a bigram model assigns a probability of:
$$p(w_1, \cdots, w_m) = p(w_1) p(w_2|w_1) \cdots p(w_m|w_{m-1})$$
to a held out document. Our model assign:
\begin{equation}
p(a_1 | s_1) \cdots p(a_t | s_t) = F(s_1, a_1) \cdots F(s_t, a_t)
\end{equation}
probability to the held out driving sequence with actions $a_1 \cdots a_t$, conditioned on world states $s_1 \cdots s_t$. We define the action predictive perplexity of our model on one held out sample as:
\begin{equation}
perplexity = \exp \Big\{ -\frac{1}{t} \sum_{i=1}^t \log F(s_i, a_i) \Big\}
\end{equation}

To evaluate a model, one can take the most probable action predicted $a_{pred} = \textrm{argmax}_{a} F(s, a) $ and compare it with the action $a_{real}$ that is carried out by the driver. This is the accuracy of the predictions from a model. Note that models generally do not achieve 100\% accuracy, since a driving model does not know the intention of the driver ahead of time. 

\subsection{Discrete and Continuous Action Prediction}
The output of our driving model is a probability distribution over all possible actions. 
A driving model should have correct motion action predictions despite encountering complicated scenes such as an intersection, traffic light, and/or pedestrians.  We first consider the case of discrete motion actions, and then investigate continuous prediction tasks, in both cases taking into account the prediction of multiple modes in a distribution  when there are multiple possible actions.

\mysubsubsection{Discrete Actions}
In the discrete case, we train our network by minimizing perplexity on the training set. In practice, this effectively becomes minimizing the cross entropy loss between our prediction and the action that is carried out. In real world of driving, it's more prevalent to go straight, compared to turn left or right. Thus the samples in the training set are highly biased toward going straight. Inspired by \cite{zhang2016colorful}, we investigated the weighted loss of different actions according to the inverse of their prevalence.

\mysubsubsection{Continuous Actions}
\label{sec:continuous_formulate}
To output a distribution in the continuous domain, one could either use a parametric approach, by defining a family of parametric distribution and regressing to the parameters of the distribution, or one can employ a non-parametric approach, e.g. discretizing the action spaces into many small bins. Here we employ the second approach, since it can be difficult to find a parametric distribution family that could fit all scenarios. 

\subsection{Driving with Privileged Information}
\label{sec:ptrain}
\begin{figure}
\begin{center}
\includegraphics[width=1\linewidth]{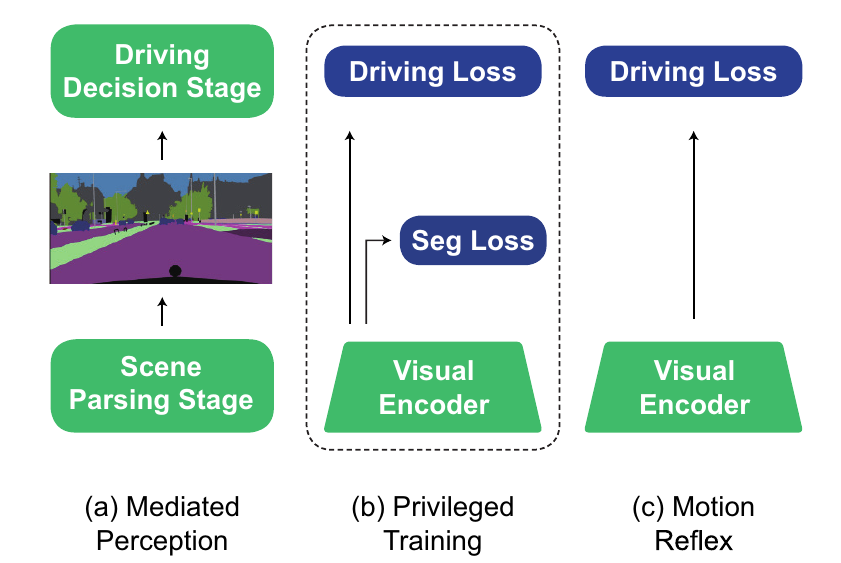}
\end{center}
\caption{Comparison of learning approaches. Mediated Perception relies on  semantic-class  labels at the pixel level alone to drive motion prediction. The Motion Reflex method learns a representation based on raw pixels. Privileged Training learns from raw pixels but allows side-training on semantic segmentation tasks. }
\label{fig:PTrain}
\vskip -0.4cm
\end{figure}

Despite the large-scale nature of our training set, small phenomena and objects may be hard to learn in a purely end-to-end fashion. We propose to exploit privileged learning \cite{vapnik2009new,sharmanska2013learning,Hoffman_CVPR2016} to learn a driving policy that exploits both task loss and available side losses.  In our model, we use semantic segmentation as the extra supervision. Figure  \ref{fig:PTrain} summarizes our approach and the alternatives: motion prediction could be learned fully end to end (Motion Reflex Approach), or could rely fully on predicted intermediate semantic segmentation labels (Mediated Perception Approach), in contrast, our proposed approach (Privileged Training Approach) adopts the best of both worlds, having the semantic segmentation as a side task to improve the representation, which ultimately performs motion prediction.  Specifically, we add a segmentation loss after fc7, which enforces fc7 to learn a meaningful feature representation. Our results below confirm that even when semantic segmentation is not the ultimate goal, learning with semantic segmentation side tasks can improve performance, especially when coercing a model to attend to small relevant scene phenomena.

\section{Dataset}
\begin{figure}
\begin{center}
\includegraphics[width=1\linewidth]{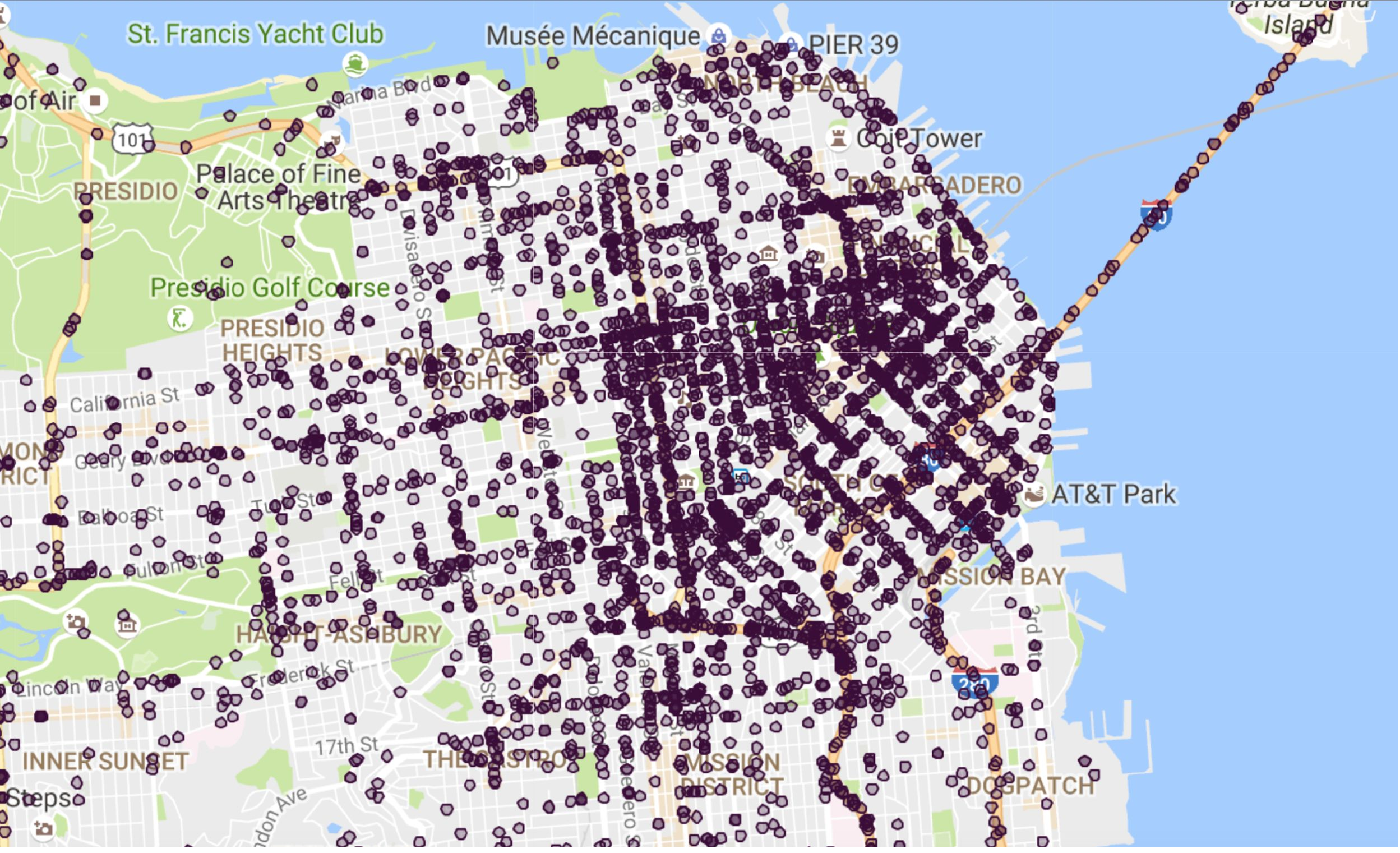}
\end{center}
\caption{Example density of data distribution of BDDV in a major city. Each dot represents the starting location of a short video clip of approximately 40 seconds.}
\label{fig:scatter}
\end{figure}

\begin{table*}[t]
\begin{center}
\small
\resizebox{\textwidth}{!}{
\begin{tabular}{|l|c|c|c|c|c|c|}
\hline
\textbf{Datasets} & settings & type & Approx scale &   Diversity & Specific Car Make\\
\hline
KITTI               &\textbf{city, rural area, highway}  &  \textbf{real}      &  less than 1 hour      &  one city, one weather condition, daytime         & Yes  \\
Cityscape           &city                            &  \textbf{real}      &    less than 100 hours    &  German cities, \textbf{multiple weather conditions}, daytime  &  Yes \\
Comma.ai            &mostly highway                  &  \textbf{real}      &    7.3 hours    &  highway, N.A. , \textbf{daytime and night}          & Yes  \\      
Oxford              &city            &  \textbf{real}      &    214 hours   &  one city (Oxford), \textbf{multiple weather conditions}, daytime        & Yes  \\
Princeton Torcs     &highway                         &  synthesis      &    13.5  hours   &  N.A.          &  N.A. \\
GTA                 &city, highway                   &  synthesis      &    N.A.          &  N.A.         &   N.A. \\
BDDV(ours)           &\textbf{city, rural area, highway}  &  \textbf{real}      &    \textbf{10k hours} & \textbf{multiple cities}, \textbf{multiple weather conditions},\textbf{daytime and night}   &  \textbf{No} \\
\hline
\end{tabular}}
\end{center}
\caption{Comparison of our dataset with other driving datasets.}
\label{table:1}
\end{table*}

The Berkeley DeepDrive Video dataset (BDDV) is a dataset comprised of real driving videos and GPS/IMU data.
The BDDV dataset contains diverse driving scenarios including cities, highways, towns, and rural areas in several major cities in US.
We analyze different properties of this dataset in the following sections and show its suitability for learning a generic driving model in comparison with sets of benchmark datasets including KITTI, Cityscapes, Comma.ai dataset, Oxford Dataset, Princeton Torcs, GTA, each of which varies in size, target, and types of data. A  comparison of datasets is provided in Table \ref{table:1}.
\subsection{Scale}
BDDV provides a collection of sufficiently large and diverse driving data, from which it is possible to learn generic driving models. The 
BDDV contains over 10,000 hours of driving dash-cam video streams from different locations in the world. 
The largest prior dataset is Robotcar dataset~\cite{RobotCarDatasetIJRR} 
which corresponds to 214 hours of driving experience. KITTI, which has diverse calibrated data, provides 22 sequences (less than an hour) for SLAM purposes. In Cityscapes, there are no more than 100 hours driving video data provided upon request. To the best of knowledge,  BDDV is at least in two orders larger than any benchmark public datasets for vision-based autonomous driving.

\subsection{Modalities}
Besides the images, our BDDV dataset also comes with sensor readings of a smart phone. The sensors are GPS, IMU, gyroscope and magnetometer. The data also comes with sensor-fused measurements, such as course and speed. Those modalities could be used to recover the trajectory and dynamics of the vehicle.

\subsection{Diversity}
\begin{figure}
\begin{center}
\includegraphics[width=1\linewidth]{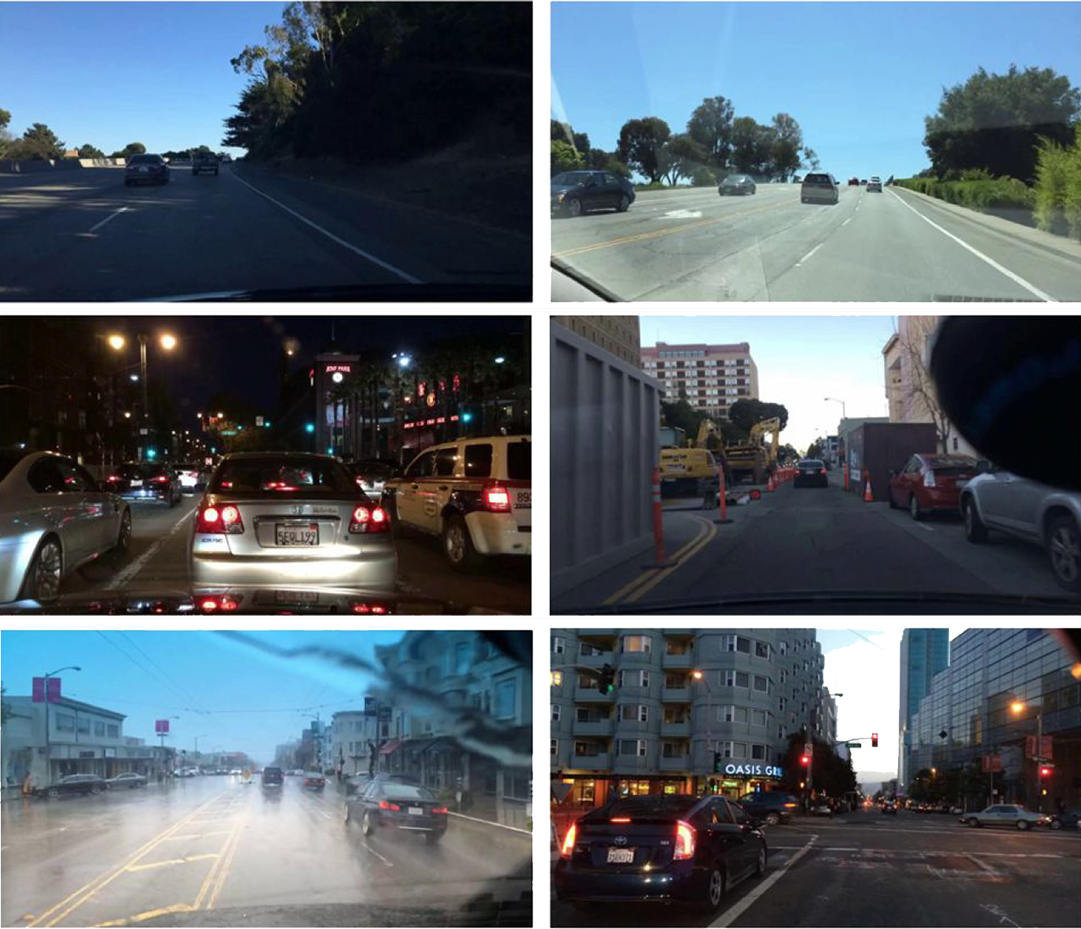}
\end{center}
\caption{Sample frames from the BDDV dataset.}
\label{fig:data_dist}
\vspace*{-2ex}
\end{figure}

The BDDV dataset is collected to learn a driving model that is generic in terms of  driving scenes, car makes and models, and driving behaviors. The coverage of  BDDV includes various driving, scene, and lighting conditions. In Figure \ref{fig:data_dist} we show some samples of our dataset in nighttime, daytime, city areas, highway and rural areas.  
As shown in Table \ref{table:1}, existing benchmark datasets are limited in the variety of scene types they comprise.
In Figure \ref{fig:scatter} we illustrate the spatial distribution of our data across a major city.

\vspace{-.3cm}
\section{Experiments}
\label{sec:experiment}
For our initial experiments, we used a subset of the BDDV comprising 21,808 dashboard camera videos as training data, 1,470 as validation data and 3,561 as test data. Each video is approximately 40 seconds in length. Since a small portion of the videos has duration just under 40 seconds, we truncate all videos to 36 seconds. We downsample  frames 
to $640\times 360$ and temporally downsample the video to 3Hz to avoid feeding near-duplicate frames into our model. After all such preprocessing, we have a total of 2.9 million frames, which is approximately 2.5 times the size of the ILSVRC2012 dataset. 
To train our model, we used stochastic gradient descent (SGD) with an initial learning rate of $10^{-4}$, momentum of 0.99 and a batch size of 2. The learning rate was decayed by 0.5 whenever the training loss plateaus. Gradient clipping of 10 was applied to avoid gradient explosion in the LSTM.
The LSTM is run sequentially on the video with the previous visual observations. Specifically, the number of hidden units in LSTM is 64.
Models are evaluated using predictive perplexity and  accuracy, where the maximum likelihood action is taken as the prediction.

\subsection{Discrete Action Driving Model}
\label{sec:discrete}

We first consider the discrete action case, in which we define four actions: \textbf{straight}, \textbf{stop}, \textbf{left turn}, \textbf{right turn}. The task is defined as predicting the feasible actions in the next $1/3$rd of a second. 

\begin{table}[t]
\begin{center}
\small
\resizebox{0.47\textwidth}{!}{
\begin{tabular}{|l|c|c|c|c|c|}
\hline
Configuration           & Image        & Temporal       & Speed         & Perplexity         & Accuracy   \\
\hline
Random-Guess         & N.A.           & N.A.             & No             & 0.989              & 42.1\%         \\
Speed-Only           & N.A.          & LSTM             & Yes            & 0.555          & 80.1\%     \\
CNN-1-Frame          & CNN           & N.A.             & No             & 0.491          & 82.0\%      \\
TCNN3                 & CNN           & CNN              & No             & 0.445          & 83.2\% \\
TCNN9                 & CNN           & CNN              & No             & 0.411          & 84.6\% \\
CNN-LSTM             & CNN           & LSTM             & No             & 0.419          & 84.5\%     \\
CNN-LSTM+Speed              & CNN           & LSTM             & Yes            & 0.449       & 84.2\%     \\
FCN-LSTM             & FCN           & LSTM             & No             & 0.430              & 84.1\%          \\
\hline
\end{tabular}}
\end{center}
\caption{Results on the discrete feasible action prediction task. We investigated the influence of various image encoders, temporal networks and the effect of speed. Log perplexity (lower is better) and accuracy (higher is better) of our prediction are reported. See Section \ref{sec:discrete} for details.}
\label{tab:discrete}
\end{table}

Following Section \ref{sec:arch}, we minimize perplexity on the training set and evaluate perplexity and accuracy of the maximum likelihood prediction on a set of held out videos. In Table \ref{tab:discrete}, we do an ablation study to investigate the importance of different components of our model. 

Table \ref{tab:discrete} shows the comparison among a few variants of our method. The Random Guess baseline  predicts randomly based on the input distribution. In the speed-only condition, we only use the speed of the previous frame as input, ignoring the image input completely. It achieves  decent performance, since the driving behavior is largely predictable from the speed in previous moment. In the ``1-Frame" configuration, we only feed in a single image at each timestep and use a CNN as the visual encoder. 
It achieves better performance than the two baseline models (random and speed-only). This is intuitive, since human drivers can get a  good, but not perfect, sense of feasible motions from a single frame. In the TCNN configuration we study using temporal convolution as the temporal fusion mechanism. We used a fixed length window of 3 (TCNN3) and 9 (TCNN9), which is 1 and 3 seconds in time respectively. TCNN models further improves the performance and the longer the time horizon, the better the performance. However, it needs a fixed size of history window and is more memory demanding than the LSTM based approach. We also explore the CNN-LSTM approach, and it achieves comparable performance as TCNN9. When changing the visual encoder from CNN to FCN, the performance is comparable. However, as we will show later \ref{sec:ptrain}, a FCN-based visual encoder is vital for learning from  privileged segmentation information. We also found that the inverse frequency weighting of the loss function \cite{zhang2016colorful} encourages the prediction of rare actions, but it does not improve the prediction perplexity. Thus we do not use this  in our methods above. 

\begin{figure}[t]
    \centering
    \begin{subfigure}[t]{0.24\textwidth}
        \centering
        \includegraphics[width=\textwidth]{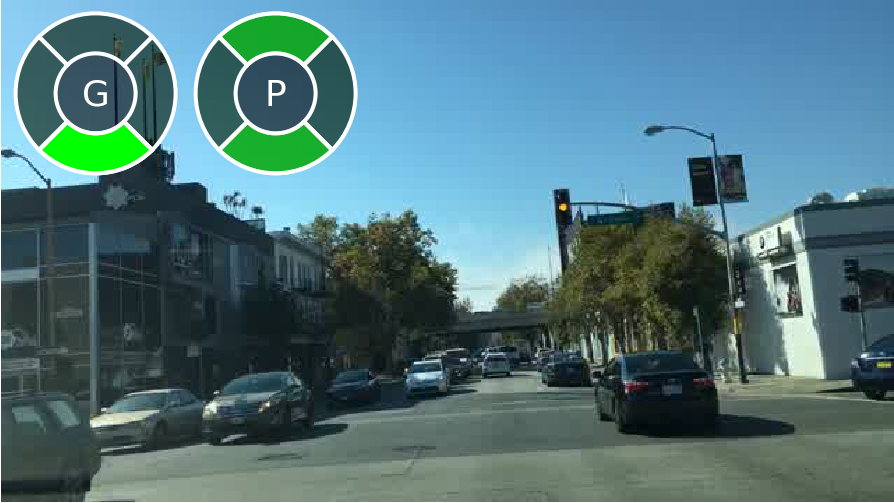}
        \caption{go at yellow light}
    \end{subfigure}%
    ~ 
    \begin{subfigure}[t]{0.24\textwidth}
        \centering
        \includegraphics[width=\textwidth]{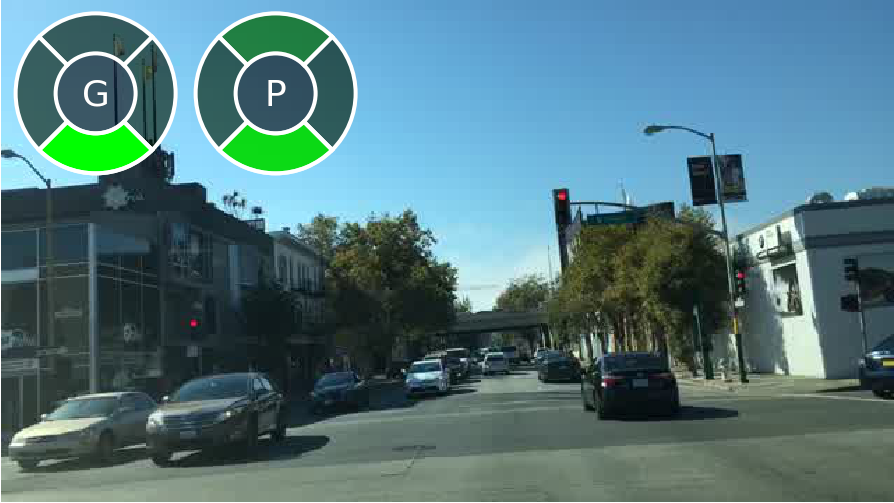}
        \caption{stop at red light}
    \end{subfigure}
    ~ 
    \begin{subfigure}[t]{0.24\textwidth}
        \centering
        \includegraphics[width=\textwidth]{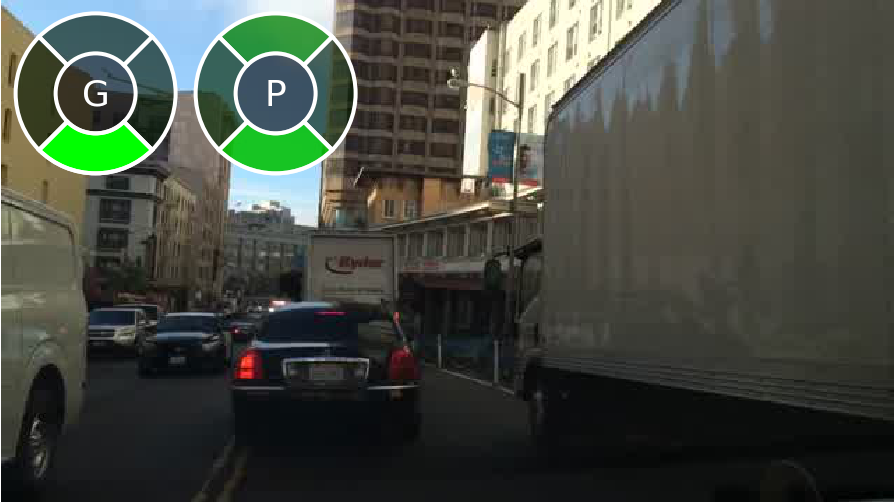}
        \caption{stop \& go equal weight at medium distance}
    \end{subfigure}%
    ~ 
    \begin{subfigure}[t]{0.24\textwidth}
        \centering
        \includegraphics[width=\textwidth]{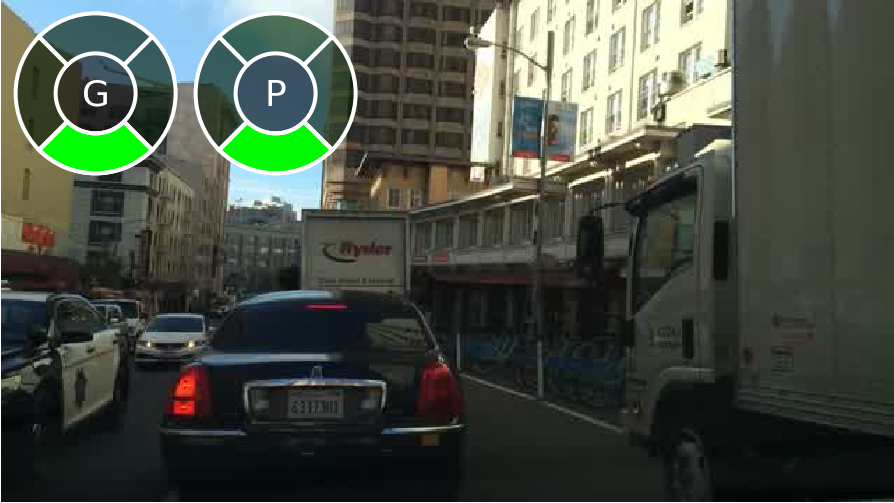}
        \caption{stop when too close to vehicle ahead}
    \end{subfigure}
    \caption{Discrete actions predicted by our FCN-LSTM model. Each row of 2 images show how the prediction changes by time. The green bars shows the probability of doing that action at that time. The red bars are the driver's action. The four actions from top to bottom are going straight, slow or stop, turn left and turn right. }
    \label{fig:discrete_prediction}
    \vskip -.5cm
\end{figure}

In Figure. \ref{fig:discrete_prediction}, we show some predictions made by our model. In the first pair of images (subfig. a\&b), the car is going through an intersection, when the traffic light starts to change from yellow to red. Our model has  predicted to go straight when the light is yellow, and the prediction changes to stop  when the traffic light is red. This indicates that our model has learned how human drivers often react to traffic light colors.  In the second pair (c\& d), the car is approaching a stopped car in the front. In (c), there is still empty space ahead, and our model predicts to go or stop roughly equally. However, when the driver moves closer to the front car, our model predicts stop instead. This shows that our model has learned the concept of distance and automatically map it to the feasible driving action.

\begin{table}[h]
\centering
\caption{Continuous lane following experiment. See Section \ref{sec:continuous} for details.}
\label{table:continuous}
\begin{tabular}{|l|l|}
\hline
Configuration    & Angle Perplexity \\
\hline
Random Guess & 1.86 \\
Linear Bins   & -2.82             \\
Log Bins      & -3.66             \\
Data-Driven Bins & -4.83             \\
\hline
\end{tabular}
\end{table}

\subsection{Continuous Action Driving Model}
\label{sec:continuous}
In this section, we investigate the continuous action prediction problem, in particular,  lane following. We define the lane following problem as predicting the angular speed of the vehicle in the future $1/3$ second. As proposed above, we discretize the prediction domain into bins and turn the problem into a multi-nomial prediction task.

We evaluated three different kinds of binning methods (Table \ref{table:continuous}). First we tried a linear binning method, where we discretize  $[-90^\circ, 90^\circ]$ into 180 bins of width $1^\circ$ . 
The linear binning method is reasonable under the assumption that constant controlling accuracy is needed to drive well.
Another reasonable assumption might be that constant relative accuracy is required to control the turns. This corresponds to the log bins method. We use a total of 180 bins that is evenly distributed in $logspace(-90^\circ, -1^\circ)$ and $logspace(1^\circ, 90^\circ)$. We also tried a data-driven approach. We first compute the distribution of the drivers' behavior (the vehicle's angular velocity) in the continuous space. Then we discretize the distribution to 180 bins, by requiring each bin having the same probability density. Such data-driven binning method will adaptively capture the details of the driver's action. During training we use a Gaussian smoothing with standard deviation of $0.5$ to smooth the training labels in nearby bins. 
Results are shown in Table \ref{table:continuous}; 
The data-driven binning method performed the best among all of them, while the linear binning performed worst.

\begin{figure}[t!]
    \centering
    \begin{subfigure}[t]{0.24\textwidth}
        \centering
        \includegraphics[width=\textwidth]{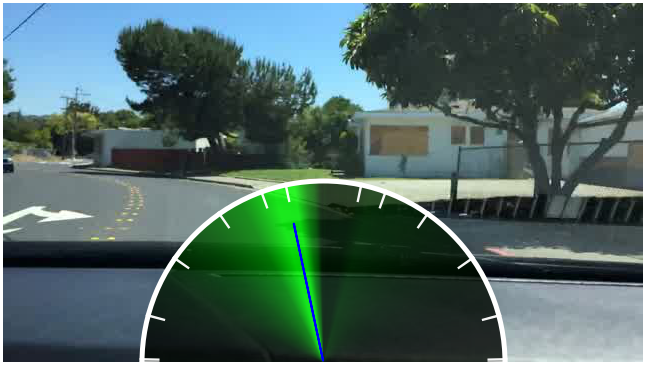}
        \caption{lane following left}
    \end{subfigure}%
    ~ 
    \begin{subfigure}[t]{0.24\textwidth}
        \centering
        \includegraphics[width=\textwidth]{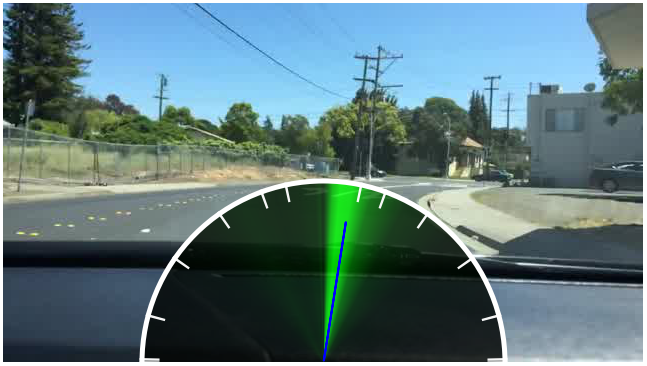}
        \caption{lane following right}
    \end{subfigure}
    ~ 
    \begin{subfigure}[t]{0.24\textwidth}
        \centering
        \includegraphics[width=\textwidth]{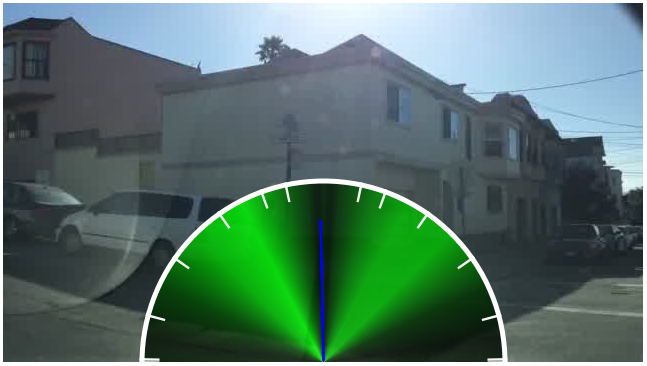}
        \caption{multiple possible actions: turn left or go straight}
    \end{subfigure}%
    ~ 
    \begin{subfigure}[t]{0.24\textwidth}
        \centering
        \includegraphics[width=\textwidth]{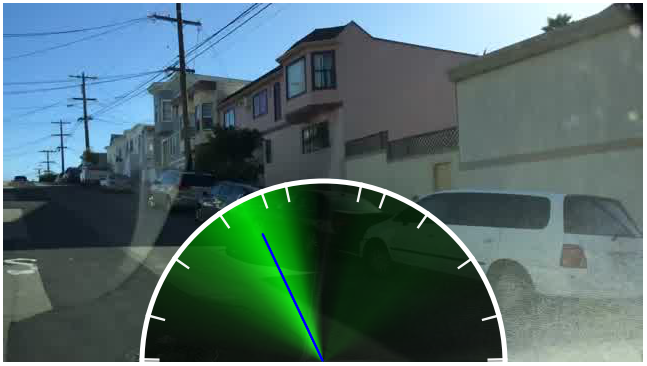}
        \caption{collapsed to single action after the turn}
    \end{subfigure}
    ~ 
    \begin{subfigure}[t]{0.24\textwidth}
        \centering
        \includegraphics[width=\textwidth]{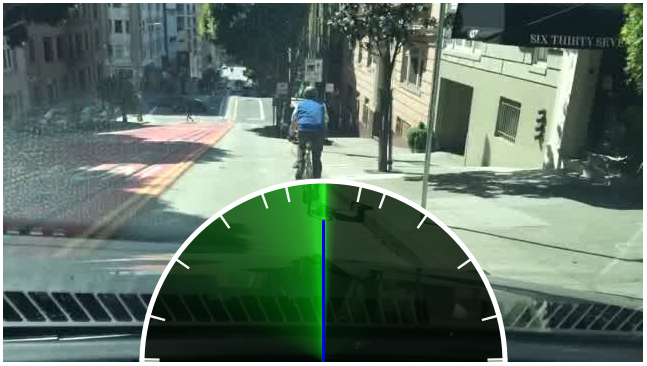}
        \caption{single sided prediction due to side walk}
    \end{subfigure}%
    ~ 
    \begin{subfigure}[t]{0.24\textwidth}
        \centering
        \includegraphics[width=\textwidth]{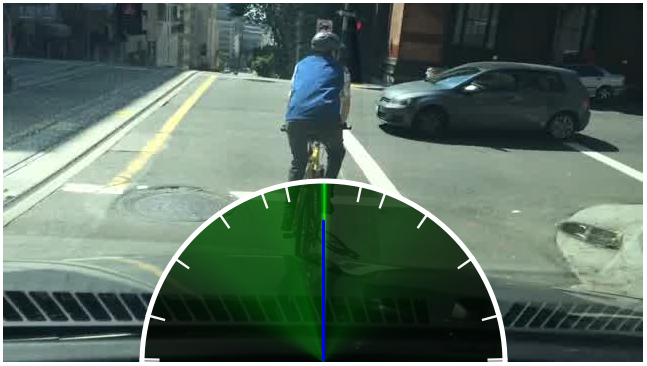}
        \caption{right turn becomes available at intersection}
    \end{subfigure}
    \caption{Continuous actions predicted by our model. The green sector with different darkness shows the probability map of going to a particular direction. The blue line shows the driver's action. }
    \label{fig:continuous_prediction}
\end{figure}

Figure \ref{fig:continuous_prediction} shows examples of our prediction on video frames. Sub-figure (a) \& (b) shows that our models could follow the curving lane  accurately. The prediction has a longer tail towards the direction of turning, which is expected since it's fine to have different degrees of turns. Sub-figure (c) shows the prediction when a car is starting to turn left at an intersection. It assigns a higher probability to continue turning left, while still assigning a small probability to go straight. The probability in the middle is close to zero, since the car should not hit the wall. Close to the completion of the turn (sub-figure (d)), the car could only finish the turn and thus the other direction disappears. This shows that we could predict a variable number of modalities appropriately. In sub-figure (e), when the car is going close to the sidewalk on its right, our model assigns zero probability to turn right. When going to the intersection, the model has correctly assigned non-zero probability to turning right, since it's clear by that time. 

\subsection{Learning with Privileged Information (LUPI)}
\def\imagetop#1{\vtop{\null\hbox{#1}}}
\begin{figure}
	\setlength{\tabcolsep}{1pt}
    \begin{tabular}{l l l}    \imagetop{(a)} & \imagetop{\includegraphics[width=.22\textwidth]{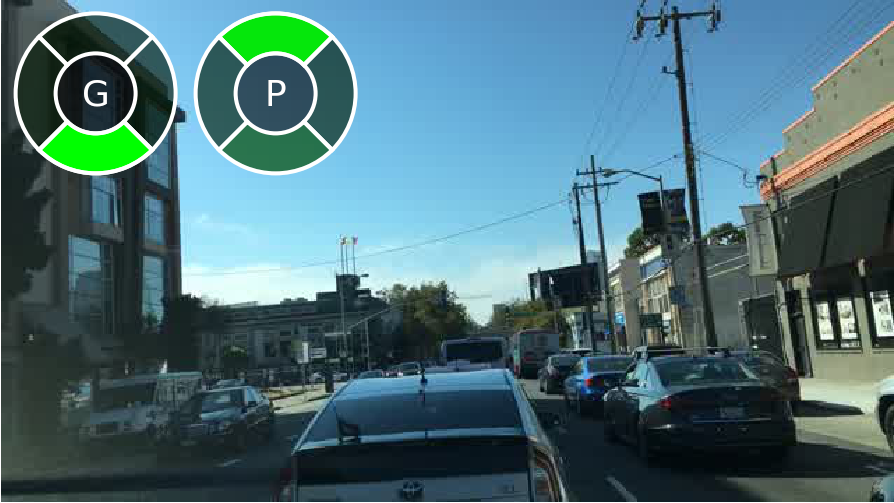}} &  \imagetop{\includegraphics[width=.22\textwidth]{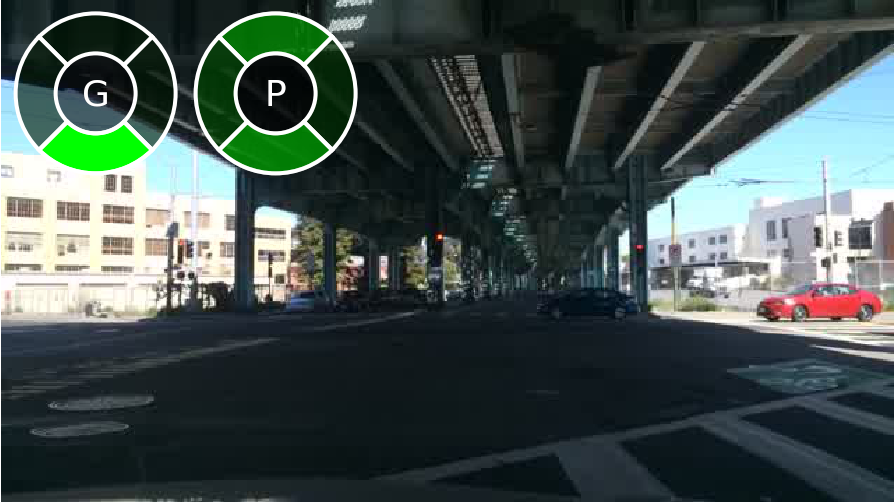}}\\[-0.2cm]
   \imagetop{(b)} & \imagetop{\includegraphics[width=.22\textwidth]{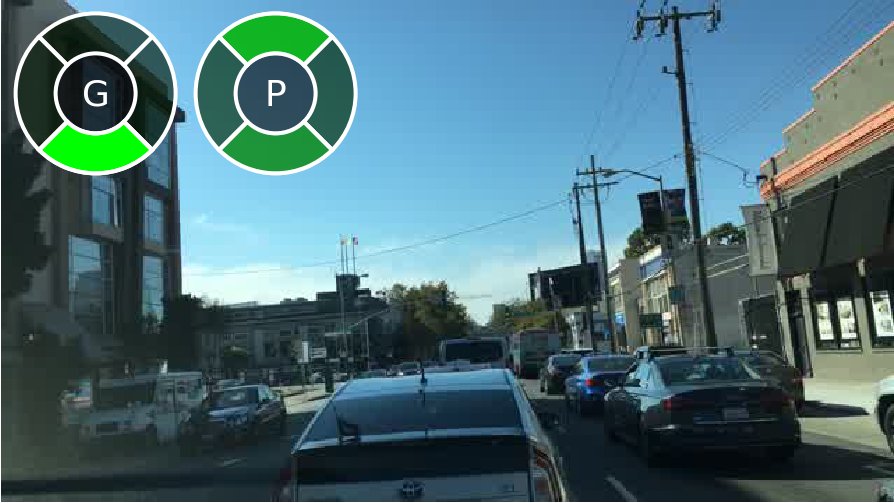}} & \imagetop{\includegraphics[width=.22\textwidth]{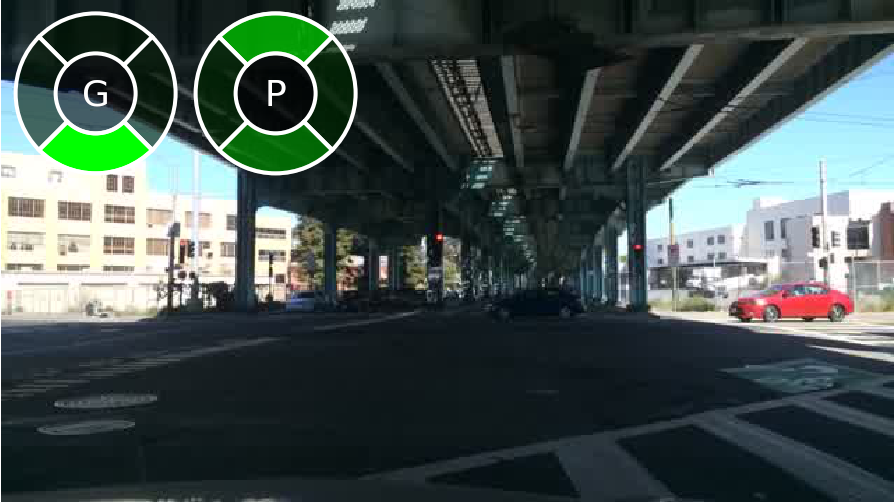}}\\[-0.2cm]
   \imagetop{(c)} & \imagetop{\includegraphics[width=.22\textwidth]{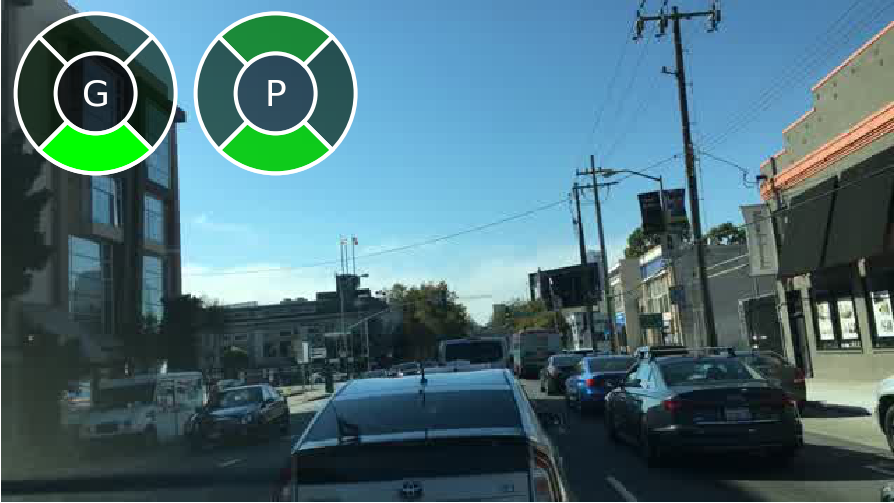}} &\imagetop{\includegraphics[width=.22\textwidth]{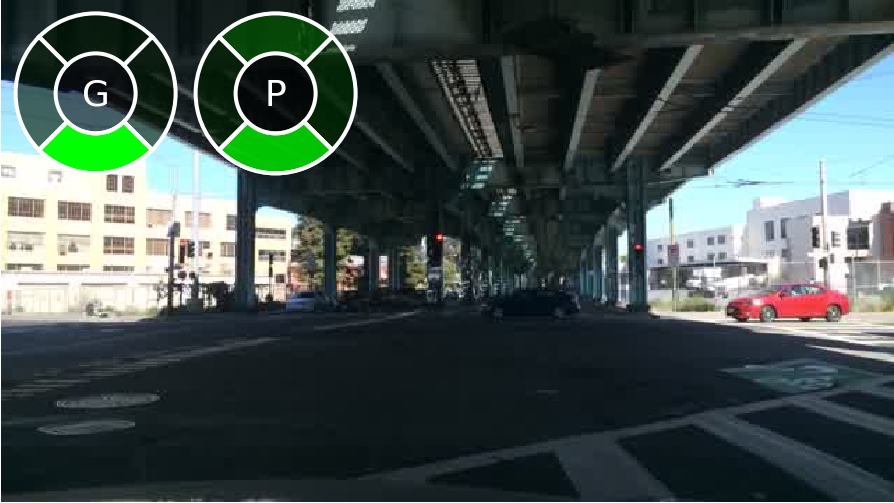}}
    \end{tabular}
    \caption{We show one example result in each column from each of the three models. (a) is the Behavior Reflex Approach. (b) is the Mediated Perception Approach and (c) the Privileged Training Approach.}
\label{fig:PT}
\vskip -0.4cm
\end{figure}

\begin{table}[t]
\begin{center}
\small
\begin{tabular}{|l|c|c|}
\hline
\textbf{method} & perplexity & accuracy\\
\hline
Motion Reflex Approach                  &        0.718                &     71.31\%   \\
Mediated Perception Approach            &        0.8887                &      61.66    \\
Privileged Training Approach            &        \textbf{0.697}       &     \textbf{72.4\%}     \\      

\hline

\end{tabular}
\end{center}
\caption{Comparison of the privileged training with other methods.}
\label{table:new_pt}
\end{table}

In this section, we demonstrate our LUPI approach on the discrete action prediction task. Following Section \ref{sec:ptrain}, we designed three approaches: The Motion Reflex Approach refers to the FCN-LSTM approach above. The Privileged Training approach takes the FCN-LSTM architecture and adds an extra segmentation loss after the fc7 layer. We used BDD Segmentation masks as the extra supervision. Since the BDDV dataset only contains the car egomotion and the BDD Segmentation dataset only contains the segmentation of individual images, we pair each video clip with 10 BDD Segmentation images during training. The motion prediction loss (or driving loss) and the semantic segmentation loss are weighted equally. For the Mediated Perception Approach, we first compute the segmentation output of every frame in the videos using the Multi-Scale Context Aggregation approach described in \cite{yu2015multi}. We then feed the segmentation results into an LSTM and train the LSTM independently from the segmentation part, mimicking stage-by-stage training. In theory, one would not need side task to improve the performance of a neural network with unlimited data. To simulate a scenario where we only have limited amount of training data, we run experiments on a common subset of 1000 video clips.

As shown in Table \ref{table:new_pt}, the Privileged Training approach achieves the best performance in both perplexity and accuracy. These observations align well with our intuition that training on side tasks in an end-to-end fashion improves performance. Figure \ref{fig:PT} shows an example in which Privileged Training provides a benefit. In the first column, there is a red light far ahead in the intersection. The Privileged Training approach has successfully identified that and predicted stop 
in (c), while the other two methods fail. In the second column, the car is waiting behind another car. In the frame immediately previous to these frames, the vehicle in front had an illuminated brake light. The second column of images shows the prediction of the three methods when the brake light of the car goes out but the vehicle has not yet started to move. The Privileged Training approach in (c) predicts 
stop 
with high probability. The other two methods behave more aggressively and predict going straight with high probability. 

\vskip -2cm
\section{Conclusion}
We introduce an approach to learning a generic driving model from large scale crowd-sourced video dataset with an end-to-end trainable architecture. It can learning from monocular camera observations and previous egomotion states to predict a distribution over future egomotion. The model uses a novel FCN-LSTM architecture to learn from driving behaviors. It can take advantage of semantic segmentation as side tasks improve performance, following the privileged learning paradigm.
To facilitate our study, we provide a novel large-scale dataset of crowd-sourced driving behaviors that is suitable for learning driving models. We investigate the effectiveness of our driving model and the ``privileged" learning by evaluating future egomotion prediction on held-out sequences across diverse conditions.
\vskip -1cm
\section*{Acknowledgement}
Prof. Darrell was supported in part by DARPA; NSF awards  IIS - 1212798, IIS - 1427425, and IIS - 1536003, Berkeley DeepDrive, and the Berkeley Artificial Intelligence Research Center. We appreciate BDD sponsor Nexar for providing the Berkeley DeepDrive Video Dataset.
{\small
\bibliographystyle{ieee}
\bibliography{egbib}
}

\end{document}